\documentclass[10pt,twocolumn,letterpaper]{article}

\usepackage{iccv}
\usepackage{times}
\usepackage{epsfig}
\usepackage{graphicx}
\usepackage{amsmath}
\usepackage{amssymb}

\usepackage{booktabs,multirow,adjustbox}
\usepackage{amsmath,amssymb,amsfonts,dsfont,bm}
\usepackage{algorithm}
\usepackage{algorithmic}
\usepackage{comment}
\usepackage{hhline}
\usepackage{diagbox}
\usepackage{multicol}
\usepackage[breaklinks=true,bookmarks=false]{hyperref}

\iccvfinalcopy 

\def\iccvPaperID{****} 

\ificcvfinal\fi

\begin{document}

\title{Fixing the Teacher-Student Knowledge Discrepancy in Distillation}

\author{Jiangfan Han$^{1 *}$, Mengya Gao$^{2}$\thanks{Equal contribution.}, Yujie Wang$^{2}$, Quanquan Li$^{2}$,  Hongsheng Li$^{1}$, Xiaogang Wang$^{1}$ \\
$^{1}$CUHK-SenseTime Joint Laboratory, The Chinese University of Hong Kong \\ $^{2}$SenseTime Research\\
{\tt\small\{jiangfanhan@link., hsli@ee., xgwang@ee.\}cuhk.edu.hk,} \\
{\tt\small\{gaomengya, wangyujie, liquanquan\}@sensetime.com }
}

\maketitle
\ificcvfinal\thispagestyle{empty}\fi

\begin{abstract}
Training a small student network with the guidance of a larger teacher network is an effective way to promote the performance of the student. 
Despite the different types, the guided knowledge used to distill is always kept unchanged for different teacher and student pairs in previous knowledge distillation methods. 
However, we find that teacher and student models with different networks or trained from different initialization could have distinct feature representations among different channels. (\eg the high activated channel for different categories).
We name this incongruous representation of channels as teacher-student knowledge discrepancy in the distillation process.
Ignoring the knowledge discrepancy problem of teacher and student models will make the learning of student from teacher more difficult. 
To solve this problem, in this paper, we propose a novel student-dependent distillation method, knowledge consistent distillation, which makes teacher's knowledge more consistent with the student and provides the best suitable knowledge to different student networks for distillation. 
Extensive experiments on different datasets (CIFAR100, ImageNet, COCO) and tasks (image classification, object detection) reveal the widely existing knowledge discrepancy problem between teachers and students and demonstrate the effectiveness of our proposed method. Our method is very flexible that can be easily combined with other state-of-the-art approaches.
\end{abstract}

\section{Introduction}


Recent years have witnessed the great success of applying convolutional neural networks (CNN) in various tasks and the remarkable performance can be owing to the deeper and wider design of network structures~\cite{szegedy2016rethinking,he2016deep,huang2017densely}.
Whereas, it is hard to deploy such heavy networks in practice where computational resource and memory space is limited.
Knowledge Distillation~\cite{hinton2014distilling} has been proposed to decrease the model size by training a small and shallow network (student) under the supervision of a larger and deeper network (teacher). 
Benefit by knowledge distillation, knowledge in a heavy teacher network can be compressed into a lightweight student network and the performance of the student can be improved. 
\begin{figure}[t]
\begin{center}
   \includegraphics[width=0.85\linewidth]{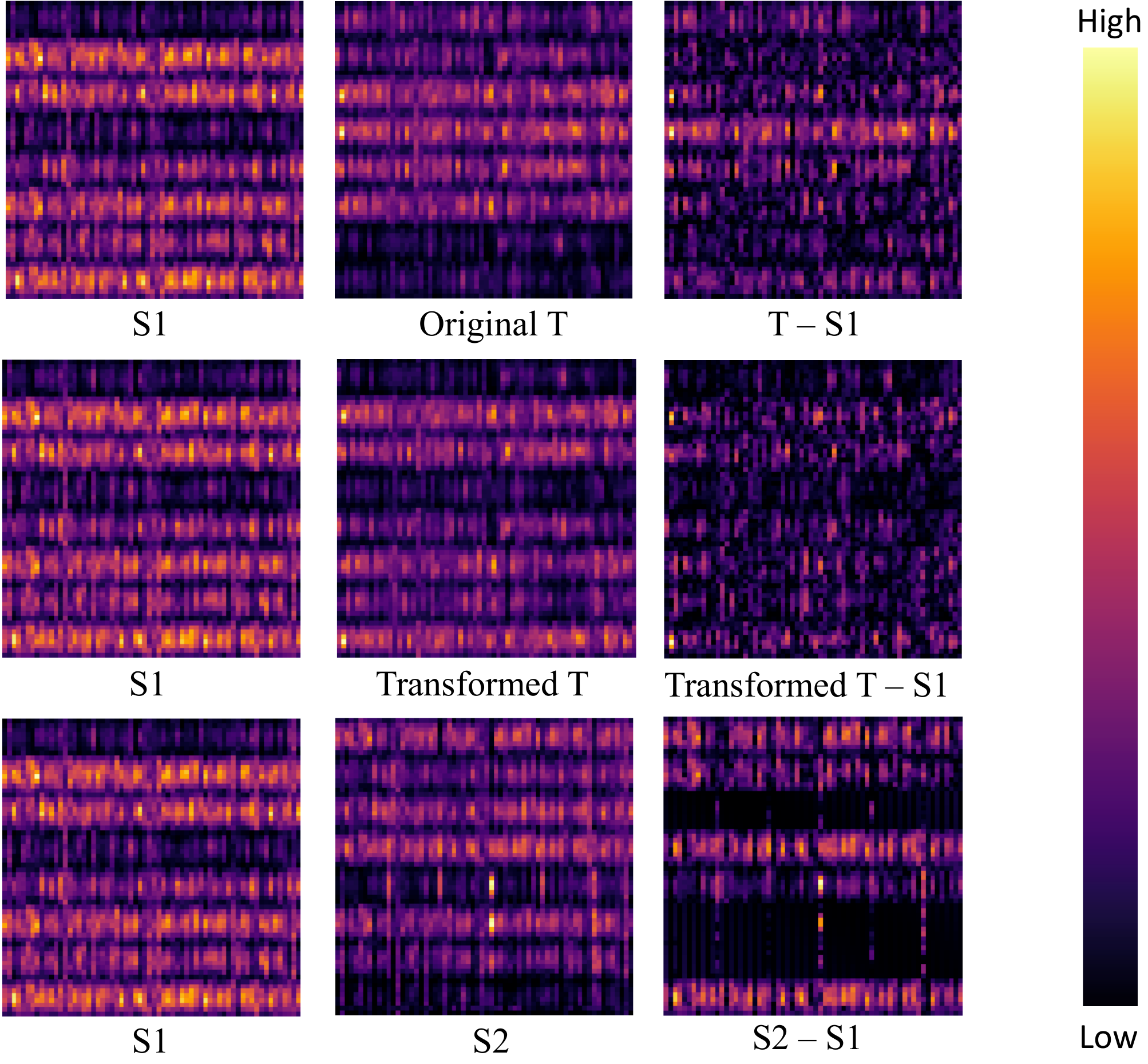}
\end{center}
\vspace{-2mm}
\caption{
   Visualization of average feature map activations of \textit{boy} category on CIFAR-100.
   \textit{S1} and \textit{S2} are the students with the same network structure ResNet20 but with different initialization.
   \textit{T} is the teacher network ResNet56.
   \textit{Transformed  T} represents the teacher's transformed feature maps by our method.
   The feature maps are extracted at the last residual block of the network.
   Each row of a subfigure illustrates an expansion of an $8\times8$ channel and there are 64 channels in total.
   The figures in the second row show the difference between the pair of average activations. Through our proposed method, the differences in activations are reduced, as shown in \textit{Transformed T - S1}.
   Best viewed in color.
   }
\label{fig:misalign}
\vspace{-4mm}
\end{figure}
Recently many works have been developed to transfer the knowledge from the teacher to the student in different ways~\cite{vaswani2017attention,tung2019similarity,liu2019structured}. 
Despite the different types of knowledge or the different ways to transfer knowledge, almost all of these works adopt the same type of knowledge between different teacher and student models.


However, the knowledge or the learned representations for the teacher and student can be distinct, as a result of different network architectures.
Recent works~\cite{frankle2018lottery,gaier2019weight} also show that initialization has a strong relationship with the final representation of the trained network.
Therefore, with the different random initialization, even models with the same network structure could have distinct feature representations.
Although most of the networks can achieve consistent performances, when analyzing the activation of intermediate layers of teacher and student models to the same sample, we find the activation responses of the same channel between different networks could have large variations, which we name such phenomenon as teacher-student knowledge discrepancy.
Visualization of knowledge discrepancy is shown in Figure~\ref{fig:misalign}. 
We gather the training data of one category and calculate the activation maps for students and teachers respectively. 
It is obvious that the activation map between T and S has a large difference.
What's more, even with the same model, \eg S1 and S2, different initialization also makes them show different activation responses.
T-S1, Transformed T-S1, S1-S2 shows the difference between activation maps.

%
%
Existing knowledge distillation methods ignore the discrepancy and directly train the student with the supervision of the teacher network.
We claim that such discrepancy will impede the learning of the student.
%
To describe the discrepancy quantitatively, we propose consistency measurements between teacher and student models. 
Based on the consistency measurements, a channel-based feature transformation is explored and deployed on the teacher feature.
The transformation can alleviate the discrepancy problem between the teacher and student dramatically and provide a suitable feature that is easier for the student to learn and obtain better performance. 
As shown in Figure~\ref{fig:misalign}, the feature of teacher model after transformation is more consistent with the student model.
Besides, our approach is independent of other knowledge distillation methods. We can integrate our method into other methods and achieve further performance improvement.

The contributions of this work can be summarized as threefold.
\begin{itemize}
    \item We demonstrate the existence of knowledge discrepancy among different teacher and student models in the knowledge distillation process.

    \item We propose a new method to solve the discrepancy problem and analyze the effectiveness of the proposed feature transformation operation.

    \item Extensive experiments on different benchmarks and visual tasks demonstrate the advantage and generalization ability of our method. Moreover, it can be easily integrated into most of the existing state-of-the-art knowledge distillation methods to further improve the performance.
\end{itemize}

\section{Related Work}

Knowledge distillation is firstly introduced in~\cite{bucilua2006model} and then brought back to popularity by~\cite{hinton2014distilling}.
The rationale behind is to use a student model ($S$) to learn from a teacher model ($T$) without sacrificing much accuracy compared to the teacher.
Challenges lie in two aspects, which are (1) how to extract the knowledge from $T$, and (2) how to transfer the knowledge to $S$.

Existing methods have designed various types of knowledge to improve their performance.
Methods in~\cite{ba2014deep} treated the hard label predicted by $T$ as the underlying knowledge, with the assumption that the well-trained $T$ has already eliminated some label errors contained in the groundtruth data.
Hinton~\etal~\cite{hinton2014distilling} argued that the soft label produced by $T$, \ie the classification probabilities, can provide richer information.
Some work~\cite{romero2015fitnets, Zagoruyko2017Paying, yim2017gift} extracted the knowledge from $T$ by processing the hidden feature map.
AT~\cite{Zagoruyko2017Paying} averaged the feature map across channel dimension to obtain spatial attention map, while work in~\cite{yim2017gift} defined inter-layer flow by computing the inner product of two feature maps, and Lee~\etal~\cite{lee2018self} improved this idea with singular value decomposition (SVD). Jin~\etal~\cite{jin2019knowledge} claimed that the representation of a converged heavy model is not easy to learn. As a result, an approach called route constrained optimization (RCO) was proposed, from the perspective of curriculum learning.

Furthermore, some methods focus on instance feature space transformation other than teaching the feature distribution of teacher to student. To address this issue, Yim~\etal~\cite{yim2017gift} proposed the Flow of Solution Procedure (FSP) which transfers the learning procedure of the teacher to the student instead of mimicking the feature of intermediate layers. Liu~\etal~\cite{liu2019relational} combined feature space transformation and instance relationship which uses input images from a training batch to calculate an instance graph and makes the student learn this graph from the teacher.

Different from all the methods above, our method focuses on how to make the knowledge distillation easier via transforming their feature maps given different pairs of teacher and student.
The proposed method can be combined with most of the knowledge distillation methods and improve their performance further.

\section{Method}
\subsection{Preliminary}
Knowledge distillation is a widely used training strategy 
which uses a well-trained teacher network giving extra guidance to the student network. Let $\mathcal{H(\theta)}$ be the network with parameter $\theta$, then $\mathcal{H}(\theta^T)$,$\mathcal{H}(\theta^{S})$ is the teacher and student network with corresponding  parameters $\theta^T$ and $\theta^S$. Note that the architecture of the teacher and student can be different.
For input samples $\mathbf{X}$ with groundtruth label $\mathbf{Y}$, we denote the activation feature map of a certain layer $k$ as $F_k = \mathcal{H}_k(\theta;\mathbf{X}) \in \mathbb{R}^{b\times c \times h \times w}$, where $b$ is the batch size of the input samples, $c$ is the number of channels, $h$ and $w$ are the height and width of the feature map. 
Then the feature map of teacher and student network should be $F_k^T = \mathcal{H}_k(\theta^T;\mathbf{X})\in \mathbb{R}^{b\times c_T \times h_T \times w_T}$ and $F_k^S = \mathcal{H}_k(\theta^S;\mathbf{X})\in \mathbb{R}^{b\times c_S \times h_S \times w_S}$  respectively. Generally speaking, the size of the feature maps between different models are not the same. But in the case of distillation, we usually use feature map with the same size, \ie we have $c_T = c_S, h_T = h_S, w_T = w_S$ and do not need to distinguish them.

For the traditional knowledge distillation methods, the training methods can be written as follows:
\begin{align}
    \mathcal{L} =  \mathcal{L}_{cls}(\mathcal{H}(\theta^S;\mathbf{X}), \mathbf{Y}) + \mathcal{L}_{dis}
\label{eq:traditional_loss}
\end{align}
\begin{align}
    \mathcal{L}_{dis} =  \sum_{(k,k') \in \Omega}{\alpha_{(k,k')}\mathcal{L}_{d}(F_k^T, F_{k'}^S)}
\label{eq:distillation_loss}
\end{align}
Where $\mathcal{L}_{cls}$ is the cross-entropy loss between the prediction of student network $\mathcal{H}(\theta^S;\mathbf{X})$ and groundtruth $\mathbf{Y}$. 
$\mathcal{L}_{dis}$ is the distillation loss measuring the difference of knowledge obtained by the teacher and learned by the student, and such distillation operation can be executed in different layers. Each loss term $\mathcal{L}_d$ can be Euclidean distance, Kullback–Leibler divergence, or other measuring metrics. $\Omega$ is the set of all teacher/student feature pairs used for knowledge distillation. $\alpha_{(k,k')}$ is the weight of the loss between the feature pair $F_k^T$ and $F_{k'}^S$.

\subsection{Teacher-Student Knowledge Discrepancy}
\label{subsec:discrepancy}
As we stated above, teacher-student knowledge discrepancy is the phenomenon that the activation responses of the same channel between the teacher and the student could have large variations.
To describe such a phenomenon quantitatively, we propose consistency measurement as following:

For the feature of teacher $F_k^T $ and student $F_{k'}^S$, a global average pooling along the height and weight dimension $\mathcal{A}$ is executed for those feature maps.
Then for teacher $\mathcal{A}(F^T_k) \in \mathbb{R}^{b\times c}$ and for student $ \mathcal{A}(F^S_{k'}) \in \mathbb{R}^{b \times c}$.
We will discuss the consistency measurement based on these features.

\textbf{$L_p$ consistency.} A natural idea of defining the consistency is using the $L_p$ norm. 
In this work, we define $L_p$ consistency, which is measured by the inverse of $L_p$ norm. The consistency matrix $M(F^T_{k}, F^S_{k'}) \in \mathbb{R}^{c\times c}$ is given by
\begin{align}
    M_{ij} = \frac{1}{||\mathcal{A}(F^T_{k})[:,i] - \mathcal{A}(F^S_{k'})[:,j]||_p} 
\label{eq:l2_similarity}
\end{align}
where $\mathcal{A}(F^T_k)[:,i] \in \mathbb{R}^b$ represent the $i$th channel of the activation feature. 

The consistency matrix $M(F^T_{k}, F^S_{k'})$ measures the consistent score for each channel pair. 
If $M_{ij}$ gets large value, which means the $ith$ channel of teacher and $jth$ channel of student get similar activation to the same sample, \ie they are consistent.
In the real case, we consider $p = 1,2$.

\textbf{Correlation consistency.} Pearson correlation coefficient, simplified as ``correlation'' can also be used as the consistency measurement. The consistency matrix $M$ under the correlation measurement is calculated by
\begin{align}
    M_{ij} = \frac{Cov(\mathcal{A}(F^T_{k})[:,i],\mathcal{A}(F^S_{k'})[:,j])}{\sigma_{\mathcal{A}(F^T_{k})[:,i]}\sigma_{\mathcal{A}(F^S_{k'})[:,j]}}
\label{eq:correlation_similarity}
\end{align}

Based on the definition of these two measurements, $L_p$ measurement will obtain a high consistency score only if two channels activate almost the same, while correlation measurement provides a looser constraint that as long as two channels have a consistent activation trend, the consistency score will be high.

There are also many other ways to define the measurement such as cosine distance and KL divergence. Their definition is quite straightforward and we will not state them in detail.

For all of the consistency measurements, larger $M_{ij}$ means the $ith$ channel of teacher and $jth$ channel are more consistent.
For most of the traditional distillation methods, they distill the knowledge straightly following the default channel order from the teacher to the student. We define the consistency score $\Gamma $ as:
\begin{align}
    \Gamma = Tr(M(F_k^T,F^S_{k'}))
\label{eq:pairing}
\end{align}
Here $Tr$ is the trace of matrix $M$. The consistency score $\Gamma $ can describe the consistency of corresponding channels of the feature maps entirely.
A larger $\Gamma $ value means they are more consistent, \ie less discrepant.
In the general case, we calculate the consistency between the trained teacher and student. Since the structure and the initialization of the teacher and student are different, their discrepancy is obvious and the consistency score $\Gamma$ is quite small in such cases. 
We believe that such an obvious discrepancy is harmful for the student to learn knowledge from the teacher. 
So we propose channel-based transformation for teacher features. The transformed feature will be more consistent with the student feature and promote the student learning procedure.



\subsection{Channel-based Feature Transformation}
\label{subsec:trans}
In this part, we will introduce the method to find the transformation $\mathcal{T}_{\theta^S_0}$. The subscript $\theta^S_0$ means the transformation is related to the structure and the initialization of the student. The transformation will be deployed on the teacher feature to reduce its discrepancy with student model. Since the transformation is defined in the channel dimension, we name this method a channel-based transformation.
To evaluate the consistency of the transformed teacher feature $\mathcal{T}_{\theta^S_0,(k,k')}(F_k^T)$ and the student feature $F^S_{k'}$. The consistency score $\Gamma $ is been modified as:
\begin{align}
    \Gamma = Tr(M(\mathcal{T}_{\theta^S_0,(k,k')}(F_k^T),F^S_{k'}))
\label{eq:pairing}
\end{align}
The traditional method can be viewed as a special case when $\mathcal{T}_{\theta^S_0}$ is an identity transformation.
Then we aim to find  $\mathcal{T}_{\theta^S_0}$ that maximizes $\Gamma$ under some constraints. We will introduce the methods in the following part.

\textbf{Greedy.}
First, we propose a straightforward solution: For each student channel, a teacher channel that has the largest consistency score to it will be reordered to match it. Assume for the $i$th channel of student, the $j$th channel of teacher obtains largest consistency score between them, \ie $j = \arg\max M[:,i]$, then the transformation is defined as 
\begin{align}
    \mathcal{T}_{\theta^S_0,(k,k')}(F_k^T)[:,i] = F_k^T[:,j]
\label{eq:greedy}
\end{align}
The greedy transformation is a 1 to N matching scenario. It can get the maximum $\Gamma$ score when each teacher channel can supervise many student channels. We call such transformation strategy as ``greedy matching'', simplified as ``greedy''. 


\textbf{Bipartite.}
Although greedy transformation can get the maximum $\Gamma$ score. 
However, a problem exists in such a greedy method. Much information contained in different channels of the teacher will be discarded in such a 1 to N matching strategy. To avoid such information loss, we propose a 1 to 1 matching to ensure we can maintain all information from the teacher. 


For the 1 to 1 matching, we want to find a transformation that contains not repeating teacher channels while maximizing $\Gamma$. The problem can be formulated as an optimal $\mathcal{T}^*$, where
\begin{align}
    \mathcal{T}^* = \ & \arg\max_\mathcal{T} \  Tr(M(\mathcal{T}_{\theta^S_0,(k,k')}(F_k^T),F^S_{k'})) \nonumber\\
   s.t. \ &\mathcal{T}_{\theta^S_0,k}(F_k^T)[:,i] = F_k^T[:,j],  
   \label{eq:bipartite}\\ 
   &j_m \ne j_n \ \text{for} \  \forall \ m\ne n \nonumber
\end{align}
If we regard the channels in teacher and student activation maps as elements in different sets and the consistency score as the weights between different elements, the solution to Equation~\ref{eq:bipartite} is actually the maximum matching of a weighted bipartite graph. 
We call such transformation as ``bipartite matching'', simplified as ``bipartite''. 
We can use a well-developed algorithm like Kuhn–Munkres algorithm to solve the problem. 
Compared with the greedy strategy, bipartite maintains all information in the teacher to supervise the learning of the student.

\textbf{Learning-based.}
Theoretically, the transformation finds by greedy and bipartite are special cases for a general linear transformation matrix, which can be represented by a Fully-Connected layer. The FC layer can also learn N to 1 matching.
What's more, a complex transformation may be learned by multi-layer networks with non-linearity.
Thus, we design two kinds of sub-networks to learn the transformation.
Sub-net-FC only contains a single Fully-Connected layer trying to learn the linear transformation of the original teacher's feature.
Sub-net-Res includes one residual block, which is used to learn non-linear transformation.

In such a scenario, Teacher's features are first fed into the sub-network, and output features $\mathcal{T}_{\theta^S_0}(F_{k}^{T})$ will be used to calculate the consistency score with corresponding students features. The consistency score will act as the loss function to optimize the parameters of sub-nets. We call such strategy ``learning-based''

\subsection{Knowledge Consistent Distillation}
\label{subsec:strategy}
%

%
Once getting the transformation $\mathcal{T}_{\theta^S_0}$, we can get a transformed teacher which is more consistent with the student. 
%

%

Then we re-initialize the student network by the same initial value $\theta^S_0$ and re-train the network from scratch with distillation. In the current stage, the distillation loss in Equation ~\ref{eq:traditional_loss} will be replaced by:
\begin{align}
    \mathcal{L}_{condis} =  \sum_{(k,k') \in \Omega}{\alpha_{(k,k')}\mathcal{L}_{d}(\mathcal{T}_{\theta^S_0,(k,k')}(F_k^T), F_{k'}^S)}
\label{eq:align_loss}
\end{align}
Where $(k,k')$ in $\mathcal{T}_{\theta^S_0,(k,k')}$ indicates the transformation for certain $(k,k')$ pair. Then the overall loss function can be written as: 
\begin{align}
    \mathcal{L} =  \mathcal{L}_{cls}(\mathcal{H}(\theta^S;\mathbf{X}), \mathbf{Y}) + \mathcal{L}_{condis}
\label{eq:align+cls}
\end{align}
where $\mathcal{L}_{condis}$ is shown in Equation~\ref{eq:align_loss}. The whole training procedure is shown in Algorithm~\ref{alg}
\begin{algorithm}[t]
\caption{Knowledge Consistent Distillation}
\label{alg}
\begin{algorithmic}[1]
\STATE Given dataset $(\mathbf{X},\mathbf{Y})$, pretrained teacher $\mathcal{H}(\theta^T)$, student $\mathcal{H}(\theta^S)$, and the initial value of student $\theta^S_0$.
\STATE Initialize student  $\theta^S \gets \theta^S_0$
\STATE Train the student on $(\mathbf{X},\mathbf{Y})$ using $\mathcal{L}_{cls}$ and get trained student $\mathcal{H}(\theta^S_1)$.
\STATE Calculate consistency matrix $M$ between $\mathcal{H}(\theta^S_1)$ and $\mathcal{H}(\theta^T)$ on the dataset $(\mathbf{X},\mathbf{Y})$
\STATE Get transformation $\mathcal{T}_{\theta^S_0}$ by one of the method proposed in Sec.\ref{subsec:trans}
\STATE Reinitialize student by the same initial value $\theta^S \gets \theta^S_0$
\STATE Train the student by Equation~\ref{eq:align+cls}
\end{algorithmic}
\end{algorithm}

\section{Experiments}
\subsection{Implementation Details}
\label{sec:appendix_exp_settings}
\textbf{CIFAR-100}~\cite{Krizhevsky2009Learning} is commonly used to validate the performance of knowledge distillation methods, which composed of 50K training images and 10K testing images from 100 classes. 
Networks of ResNet~\cite{he2016deep} family are used on this dataset.
We set the weight decay to 5e-4 and use SGD with momentum 0.9. 
The models are trained with 2 GPUs with batch size 64 on each of them.
We set the initial learning rate to be 0.1 and divide the learning rate by 10 at 32K, 48K iterations, terminating at 64K iterations.

\textbf{ImageNet}~\cite{Deng2009ImageNet} is a large-scale dataset which contains over 1M training images and 50K test images collecting from 1000 categories.
We adopt ResNet and MobileNet V2 to validate the effectiveness.
For ResNet, we set the initial learning rate to 0.2 and drop it by 0.1 at 20K, 40K, 60K iterations and terminate at 65K iterations. 
Weight decay is set to 1e-4 and batch size is 2048 with 32 GPUs.
As for MobileNetV2, the learning rate starts at 0.1 and drops by 0.1 at 45K, 75K, and 100K iterations with 130K iterations in total. 

\textbf{COCO}~\cite{Lin2014Microsoft} is a very challenging object detection benchmark that has 80 object categories.
The union of 80K train images and a 35K subset of validation images are used as the training set and 5K subset of validation images (minival) is used as the evaluation set following~\cite{Lin2017Feature,lin2017focal}.
We use standard Average Precision(AP) for evaluation and adopt FPN~\cite{Lin2017Feature} as the detection method.
The $2\times$ setting released by Detectron\footnote{https://github.com/facebookresearch/Detectron} is used.

We compare our proposed method with a number of state-of-the-art knowledge distillation methods including KD~\cite{hinton2014distilling}, FitNets~\cite{romero2015fitnets}, AT~\cite{Zagoruyko2017Paying}, AB~\cite{heo2019knowledge}, RKD~\cite{park2019relational} and RCO~\cite{jin2019knowledge}. 
For KD, we set $T=4$ and $\lambda=16, 9$ for CIFAR-100 and ImageNet, following~\cite{hinton2014distilling}.
For FitNets, AT, AB, RKD, and RCO, the loss weight is set to $10^2, 10^3, 10, 10, 10^2$, following~\cite{Zagoruyko2017Paying}\cite{jin2019knowledge}\cite{heo2019knowledge}.

To further demonstrate the generality of our approach, we also apply it together with FitNets, RKD, AB, and RCO. For FitNets, AB and RCO, they are operated on the original feature maps so we just have to apply our transformation before calculating their distillation losses. For RKD, its distillation loss is calculated among instance distances into a batch, we add another loss calculated with $L_2$ distances on transformed feature maps to combine it with our approach.

On COCO, the loss weight we used for $L_2$ loss is set to be $10^{-4}$ to balance with task loss.
%
And following~\cite{heo2019overhaul}, we apply distillation methods before ReLU at the end of the last layer block. 
%

\subsection{Ablation Studies}\label{subsec:configuration-comparison}

\noindent \textbf{Effectiveness of Different Transformation Strategies.} In Section~\ref{subsec:trans}, we propose bipartite, greedy, and learning-based strategy for transforming features from teacher to student. A randomly generated transformation is also applied to show how much these transformation strategies contribute to the final performance. Table \ref{tab:mapping-strategy} shows the comparison results of different strategies. Bipartite achieves the best performance, significantly surpasses the baseline and random transformation.

Sub-net-FC and Sub-net-Res are learning-based strategies described in Section \ref{subsec:trans}.
Intuitively, learning a transformation will achieve better results than the bipartite or greedy solution. 
However, results show that the learned combinations of teacher's features as guidance to the student will bring a large performance drop even worse than a random transformation result. 
This is because the classification ability of the learned features is not guaranteed.
Even classification supervision is added to the learning strategy, it's still hard to balance the trade-off between classification ability and transformation.
As a result, bipartite and greedy strategy are better solutions than the learning-based strategy and random strategies.

%

\noindent \textbf{Effectiveness of Different Consistency Measurements.} In Section~\ref{subsec:discrepancy}, $L_p$, correlation, KL divergence, and cosine consistency have been mentioned to measure consistency between two feature maps. Table \ref{tab:mapping-metric} shows the results of using these different measurements.
Correlation consistency measurement shows better performance than other consistency measurement selections. 
%
%


\setlength{\tabcolsep}{10pt}
\begin{table}
\begin{center}
\begin{tabular}{lc}
\toprule
\textbf{Transformation Strategy} & \textbf{Top-1} \\
\midrule
Baseline & 67.52 \\
Random Transformation & 67.57 \\
Bipartite & \textbf{68.35} \\
Greedy & 67.83 \\
Sub-net-FC & 65.47 \\
Sub-net-Res & 65.26 \\
\bottomrule
\end{tabular}
\end{center}
\caption{Ablation on different transformation strategies using correlation metric on ImageNet with MobileNetV2-0.5$\times$ as the student and ResNet50 as the teacher. The baseline is standard training with distillation but without transformation.}
\label{tab:mapping-strategy}
\end{table}

\setlength{\tabcolsep}{10pt}
\begin{table}
\begin{center}
\begin{tabular}{lc}
\toprule
\textbf{Consistency Metric} & \textbf{Top-1} \\
\midrule
$L_1$ & 68.14 \\
$L_2$ & 68.29 \\
Cosine & 67.94 \\
Correlation & \textbf{68.35} \\
KL Divergence & 68.33 \\
\bottomrule
\end{tabular}
\end{center}
\vspace{-1mm}
\caption{Ablation on different consistency metrics using bipartite transformation on ImageNet with MobileNetV2-0.5$\times$ as the student and ResNet50 as the teacher.}
\label{tab:mapping-metric}
\end{table}

\noindent \textbf{Influence of Initialization.} As shown in Algorithm \ref{alg}, we should use the same parameters $\theta_{0}^{S}$ to initialize the student while calculating transformation $\mathcal{T}_{\theta^S_0}$ and distilling student model. 
%
%
To measure the importance of keeping initialization the same, we retrain the student model with an achieved transformation matrix but different initialization.
The student model is randomly reinitialized three times, obtaining \textit{S1}, \textit{S2}, \textit{S3} with initial parameters $\theta_{1}^{S}$, $\theta_{2}^{S}$, $\theta_{3}^{S}$. 
Then, train these three student models will be trained under the guidance of the teacher using the same transform  $\mathcal{T}_{\theta^S_0}$.

As shown in Table~\ref{tab:initialization} (a), students with different initial parameters will achieve much worse performance than that with the same initial parameters $\theta_{0}^{S}$, which means that useful information is lost with randomly initialization and the effectiveness of transformation is not guaranteed under this situation.
Similarly, experiments on teacher models with different initialization can be seen in Table~\ref{tab:initialization} (b).
%
%

Results show that there is a correspondence between the achieved $\mathcal{T}_{\theta^S_0}$ and the initialization of student as well as teacher model. They prove that the initial weights will decide or at least have an apparent effect on the final feature distribution. Analysis in~\cite{frankle2018lottery} also shows the importance of initialization which guesses that initial weights are closely related to their final values after training. 

\noindent \textbf{Consistency Measurement on Random Initialized Models.} In Section~\ref{subsec:discrepancy}, the consistency is calculated on trained models. We can also calculate the consistency on random initialized models directly.
Experiments showed in Table~\ref{tab:init-mapping} use initialized student/trained teacher, trained student/trained teacher, initialized student/initialized teacher, trained student/initialized teacher four pairs to calculate the consistency respectively. Results show that with initialized student and teacher, the transformation matrix does not work well, which implies that initial weights can not replace their final trained values.

\setlength{\tabcolsep}{8pt}
\begin{table}
\small
\begin{center}
\begin{tabular}{llcccc}
    \toprule
    \multirow{3}{*}{(a)}&  &\textbf{S} & \textbf{S1} & \textbf{S2} & \textbf{S3}\\
    \cmidrule(lr){2-6}
    & \textbf{Init. Params.} & $\theta_{0}^{S}$ & $\theta_{1}^{S}$ & $\theta_{2}^{S}$ & $\theta_{3}^{S}$ \\
    \cmidrule(lr){2-6}
    & \textbf{Top-1 acc} & \bf{68.35} & 67.55 & 67.66 & 67.49\\
    \midrule
    \multirow{3}{*}{(b)}&  &\textbf{T} & \textbf{T1} & \textbf{T2} & \textbf{T3}\\
    \cmidrule(lr){2-6}
    & \textbf{Init. Params.} & $\theta_{0}^{T}$ & $\theta_{1}^{T}$ & $\theta_{2}^{T}$ & $\theta_{3}^{T}$ \\
    \cmidrule(lr){2-6}
    & \textbf{Top-1 acc} & \bf{68.35} & 67.43 & 67.62 & 67.58\\
    \bottomrule
\end{tabular}
\end{center}
\vspace{-1mm}
\caption{Ablation on relation between initialization and transformation. The effect of randomly initialized student (MobileNetV2-0.5$\times$) and teacher  (ResNet-50) is shown in (a) and (b) respectively.}
\label{tab:initialization}
\end{table}

\setlength{\tabcolsep}{15pt}
\begin{table}
\begin{center}
\small
\begin{tabular}{lcc}
    \toprule
    &\textbf{$\bm{T_{0}}$} & \textbf{$\bm{T}$}\\
    \midrule
    \textbf{$\bm{S_{0}}$} & 67.47 & 67.6\\
    \textbf{$\bm{S}$} & 67.56 & \bf{68.35}\\
    \bottomrule
\end{tabular}
\end{center}
\vspace{-1mm}
\caption{Ablation on effect of initialization in calculating transformation matrix $\mathcal{T}_{\theta^S_0}$. $S_{0}$ and $T_{0}$ refer to MobileNetV2-0.5$\times$ and ResNet50 with initial weights. S and T denote well-trained models on ImageNet.}
\label{tab:init-mapping}
\end{table}


\noindent \textbf{Comparison of Static and Dynamic Transformation.} 
As described in Section \ref{subsec:strategy} and Algorithm \ref{alg}, we seek for transformation $\mathcal{T}_{\theta^S_0}$ to alleviate discrepancy between student and teacher and $\mathcal{T}_{\theta^S_0}$ is fixed during the whole distillation process. Since the training process is dynamic, we apply experiments of dynamic transformation to show whether it is better to update the $\mathcal{T}_{\theta^S_0}$ as the training goes on. Feature of the student will be recorded in the distillation process, and $\mathcal{T}_{\theta^S_0}$ is updated every 5 epoch with the help of recorded features. 
However, experiment results in Table \ref{tab:dynamic-mapping} show that a dynamic transformation performs~(68.31\%) nearly the same as a static one~(68.35\%).

According to the hypothesis in \cite{frankle2019lottery} and ablation in the previous part, the initialization has a large influence on the final network. 
Since we use the same initialization in the student learning phase, the training process will tend to a specific result which can be represented with a static transformation.
From this perspective, a static transformation has the ability to match the features well. 
As a result, it's unnecessary to use a dynamic transformation.

\setlength{\tabcolsep}{10pt}
\begin{table}
\small
\begin{center}
\begin{tabular}{lccc}
\toprule
\textbf{Training Strategy} & \textbf{Baseline} & \textbf{Static} & \textbf{Dynamic}\\
    \midrule
    \textbf{Top-1} & 67.52 & \textbf{68.35} & 68.31\\
    \bottomrule
\end{tabular}
\end{center}
\vspace{-1mm}
\caption{Ablation on different training strategy using correlation metric and bipartite transformation on ImageNet with MobileNetV2-0.5$\times$ as student and ResNet-50 as teacher.}
\label{tab:dynamic-mapping}
\end{table}

\noindent \textbf{Number of Transformations.} 
\label{sec:appendix_intra_class_feature_alignment}
%
Note that the proposed method seeks one transformation that maps teacher features close to student features and benefits its learning. 
%
It is also possible to calculate multiple transformation matrices according to different input data, \ie use different transformations for different categories. 
In this section, we conduct experiments to compare one transformation with multiple transformations, according to the data category.

Taking ImageNet as an example, there are 1000 categories. 
We divide the data into $K$ equal-sized partitions according to its class label.
For each part, we can obtain a transformation $\mathcal{T}_{\theta^S_0}^{k}$. 
For instance, when $K=100$, the first part contains data with class label 1-10, the second contains 11-20, \textit{etc}.
The extreme situation is $K=1000$ with one class data in each part which means we calculate a transformation for each class.
Compared with the original $\mathcal{T}_{\theta^S_0}$, the difference is that only part of the data is used when calculating each transformation $\mathcal{T}_{\theta^S_0}^{k}$. 
Then for applying transformation at the student training phase, for each input image, we first figure out the part number $k$ it belongs to and find out its corresponding $\mathcal{T}_{\theta^S_0}^{k}$.
After that, $\mathcal{T}_{\theta^S_0}^{k}$ is used to transform the teacher feature and the transformed teacher feature will be employed to the knowledge distillation method.

Table~\ref{tab:class-mapping} exhibits the results. 
From the table, we can see that using more transformations leads to a slight improvement in performance, and our method already performs well compared with the extreme $K=1000$ situation.

\setlength{\tabcolsep}{8pt}
\begin{table}
\begin{center}
\small
\begin{tabular}{lccccc}
    \toprule
    \textbf{K} & \textbf{0} & \textbf{1} & \textbf{10} & \textbf{100} & \textbf{1000}\\
    \midrule
    \textbf{Top-1} & 67.52 & 68.35 & 68.32 & 68.34 & 68.39 \\ 
    \bottomrule
\end{tabular}
\end{center}
\vspace{-2mm}
\caption{Ablation on the influence of using different numbers of transformation. $K=0$ means no transforming performed which is the common knowledge distillation with $L_2$ loss. Our proposed method uses one transformation matrix which is $K=1$.}
\label{tab:class-mapping}
\end{table}

\subsection{Comparison with State-of-the-art Methods}\label{subsec:generality}
We employ our transformation on several datasets of different vision tasks to validate its advantages and generalization ability.
The implementation of knowledge distillation methods are from their source code except FitNets\cite{romero2015fitnets}, and we reproduced FitNets for comparison.
%
%

\setlength{\tabcolsep}{10pt}
\begin{table}
\small
\begin{center}
\begin{tabular}{c|cc}
    \toprule
    \diagbox{\textbf{Method}}{\textbf{Student}} & \textbf{ResNet20-0.5$\times$} & \textbf{ResNet20}\\
    \hline
    Baseline & 59.41 & 67.96 \\
    KD~\cite{hinton2014distilling} & 60.25 & 68.88\\
    FitNets~\cite{romero2015fitnets} & 60.63 & 69.09 \\
    AT~\cite{Zagoruyko2017Paying} & 60.58 & 69.42\\
    RKD~\cite{park2019relational} & 61.43 & 69.87 \\
    AB~\cite{heo2019knowledge} & 61.36 & 69.53 \\
    RCO~\cite{jin2019knowledge} & 60.87 & 69.26\\
    \hline
    Ours & 61.38 & 69.69 \\
    Ours + FitNets & 61.59 & 69.77\\
    Ours + AB & 61.87 & 70.03 \\
    Ours + RKD & \textbf{61.92} & 69.86 \\
    Ours + RCO & 61.47 & \textbf{70.13}\\
    \bottomrule
\end{tabular}
\end{center}
\vspace{-2mm}
\caption{Comparison with other knowledge distillation methods of image classification task on CIFAR-100. The teacher is ResNet-56 with 71.21\% accuracy. Baseline represents the individually trained students' performance.}
\label{tab:cifar-100}
\end{table}


\textbf{CIFAR-100.}
A ResNet56 is adopted as the teacher network while a ResNet20 and a ResNet20-0.5$\times$ as student networks. The ResNet20-0.5$\times$ is obtained by reducing the number of channels of the ResNet20 by half.
Except for that Hinton~\etal~\cite{hinton2014distilling} uses output predictions to calculate distillation loss, other methods use the output feature maps of the third residual block to calculate mimicking loss. 
We show the top-1 accuracy of each method and our proposed results in Table~\ref{tab:cifar-100}. 
From the results, we can find that our method significantly improves performance. 
To be specific, with the ResNet56 as the teacher and the ResNet20-0.5$\times$ as the student, combined with RKD, our transformation achieves top-1 accuracy of 61.92\%, improves 0.49\% compared with original RKD performance. 
For the ResNet20 as the student, the best performance achieves 70.13\% applied with RCO, which improves 2.17\% compared with the baseline.

\textbf{ImageNet.}
We perform experiments on ImageNet to validate the existence and effectiveness of feature transformation on large scale classification dataset.
Furthermore, we also adopt different architectures of teacher and student to validate them.
A ResNet50 is chosen to be the teacher while a ResNet18 and a MobileNetV2-0.5$\times$ are used to be students separately.
We compare the top-1 accuracy with other distillation method and show the results in Table~\ref{tab:imagenet}.
The results show that knowledge discrepancy also exists in large scale classification, and equipped with our method can also further improve the performances of state-of-the-art methods~\cite{jin2019knowledge}\cite{heo2019knowledge}\cite{park2019relational}.
\begin{table}
\small
\begin{center}
\begin{tabular}{c|cc}
    \toprule
    \diagbox{\textbf{Method}}{\textbf{Student}} & \textbf{ResNet18} & \textbf{MBV2-0.5$\times$}\\
    \hline
    Baseline & 70.29 & 64.27 \\
    KD~\cite{hinton2014distilling} & 70.76& 66.75\\
    FitNets~\cite{romero2015fitnets} & 70.73 & 67.58 \\
    AT~\cite{Zagoruyko2017Paying} & 70.82 & 67.66\\
    RKD~\cite{park2019relational} & 71.07 & 68.24\\
    AB~\cite{heo2019knowledge} & 71.26 & 68.49\\
    RCO~\cite{jin2019knowledge} & 71.04 & 68.21\\
    \hline
    Ours & 71.41 & 68.35 \\
    Ours + FitNets & 71.43 & 68.36\\
    Ours + AB &  71.52 & 68.62\\
    Ours + RKD & 71.27 & \textbf{68.93}\\
    Ours + RCO & \textbf{71.58} & 68.73\\
    \bottomrule
\end{tabular}
\end{center}
\vspace{-2mm}
\caption{Comparison with other knowledge distillation methods of image classification task on ImageNet. The teacher is a ResNet50 with top-1 accuracy 75.49\%. Baseline represents the individually trained students' performance.}
\label{tab:imagenet}
\end{table}

\subsection{Extended Experiments on Object Detection}
\textbf{COCO.}
To further validate the generalization ability of the proposed method on different tasks, we conducted experiments on the challenging COCO object detection benchmark.
Experiments are performed on FPN~\cite{Lin2017Feature} with backbone ResNet18 and ResNet50 as students separately.
The teacher is an FPN with backbone ResNet152.
Knowledge distillation is applied to the output of the last residual feature map of the backbone network and $L_2$ loss is used as the distillation method.
Performance has been shown in Table \ref{tab:detection}. 
All the results state that the knowledge discrepancy also exists in distillation of object detection tasks.
Our method shows a consistent performance improvement on the detection task. 
In the case of ResNet18, the results increases from 33.6 to 36.8 on AP which makes ResNet18 achieve comparable performance with ResNet50. 
Experiments on a larger student ResNet50 also shows a significant improvement from 37.7 to 40.8, outperforming the baseline by 3.1 points on AP.

\setlength{\tabcolsep}{15pt}
\begin{table}
\begin{center}
\begin{tabular}{lll}
    \toprule
    \textbf{Backbone} & \textbf{Setting} & \textbf{AP}\\
    \midrule
    ResNet-18 & Student & 33.6 \\
    ResNet-50 & Student & 37.7 \\
    ResNet-152 & Teacher & 41.6 \\
    \midrule
    ResNet-18 & $L_2$ & 36.3\\
    ResNet-18 & Ours & \textbf{36.8} \\
    \midrule
    ResNet-50 & $L_2$ & 40.4 \\
    ResNet-50 & Ours& \textbf{40.8} \\
    \bottomrule
\end{tabular}
\end{center}
\vspace{-2mm}
\caption{Object detection results on COCO benchmark with FPN. Results are described in Average Precision (AP). Last four rows use ResNet-152 as the teacher}
\label{tab:detection}
\end{table}

\section{Analysis}
\subsection{Effects on Increasing Channel Overlap}
In this section, we apply quantitative analysis on channel activations to show the discrepancy between T and S.

\noindent\textbf{Settings.} ImageNet has 1000 classes and we use the training data for each class to calculate averaging activation maps respectively.
For each class, we sort channels by the value of averaged activations.
Then, we gather the top-k sets of channel indexes for each class.
After that, for a pair of student and teacher models, the overlapping ratio for each class is calculated by the intersection of top-k sets. 
After that, the overall overlapping ratio is obtained by averaging overlapping ratios across all classes.
Table~\ref{tab:channel-overlap} shows the result of channel overlap percentage over all channels with top-10 to top-100 highest activation channels respectively. 

\noindent\textbf{Analysis.}
%
The first row in Table~\ref{tab:channel-overlap} indicates that the activation responses of student and teacher models have large variations that the highly activated channels share different indexes in the whole feature map.
%
The result is consistent with what we demonstrate in Section~\ref{subsec:discrepancy}.

We measure the overall overlapping ratio between two differently initialized students and their activation maps also show a large difference as shown in the second row of Table~\ref{tab:channel-overlap}, which implies that the initialization will have an important effect on the distribution of the activation area of models even they share the same network structure.
%
%
%
%

Our method aims to reduce this discrepancy of activation responses.
%
As depicted in the third row of Table~\ref{tab:channel-overlap}, with S and transformed T, the discrepancy is narrowed down and it will be much easier for S to learn knowledge from T.
%
%

\setlength{\tabcolsep}{8pt}
\begin{table}
\begin{center}
\small
\begin{tabular}{llcccc}
    \toprule
    \multicolumn{2}{l}{\textbf{Models}} & \textbf{Top-10} & \textbf{Top-20} & \textbf{Top-50} & \textbf{Top-100}\\
    \midrule
    \textbf{$\bm{S}$} & \textbf{$\bm{T}$} & 18\% & 24\% & 31\% & 34\% \\
    \textbf{$\bm{S_{1}}$} & \textbf{$\bm{S_{2}}$} & 32\% & 35\% & 44\% & 48\% \\
    \textbf{$\bm{S}$} & \textbf{$\bm{\mathcal{T}(T)}$} & 74\% & 78\% & 89\% & 93\% \\
    \bottomrule
\end{tabular}
\end{center}
\vspace{-2mm}
\caption{Analysis on whether channels with high activation in different models share order overlap. S and T are MobileNetV2-0.5$\times$ and ResNet50 trained on ImageNet. $S_{1}$ and $S_{2}$ are trained student with different initialization. $\mathcal{T}(T)$ refers to transformed T with the proposed method. All the results are averaged over all the 1000 classes activation maps.}
\label{tab:channel-overlap}
\end{table}

\subsection{Effects on Reducing Discrepancy}
\setlength{\tabcolsep}{8pt}
\begin{table}
\begin{center}
\small
\begin{tabular}{lccccc}
    \toprule
    \textbf{Method} & \textbf{KL divergence} & \textbf{$L_2$ distance}\\
    \midrule
    Baseline & 0.89 & 277.19 \\
    Baseline + Transformation & 0.86 & 189.93 \\
    \midrule
    FitNets~\cite{Zagoruyko2017Paying} & 0.76 & 53.74\\
    Ours + FitNets & 0.71 & 45.72\\
    \midrule   
    RKD~\cite{park2019relational} & 0.82 & 52.19\\
    Ours + RKD & 0.64 & 39.77\\
    \midrule
    AB~\cite{heo2019knowledge} & 0.73 & 47.82\\
    Ours + AB & 0.58 & 41.84\\
    \midrule
    RCO~\cite{jin2019knowledge} & 0.69 & 46.62\\
    Ours + RCO & 0.65 & 40.03\\
    \bottomrule
\end{tabular}
\end{center}
\vspace{-2mm}
\caption{Analysis on feature consistency between teacher and student.
Results are averaged over all test images on ImageNet.
\textit{Baseline} is calculated between T and trained S.
}
\label{tab:distance}
\end{table}
To see how well the knowledge distillation results improved by our proposed method, we measure the consistency of the teacher's and student's output. On the evaluation set of ImageNet, we compare the KL divergence and $L_2$ distance between the teacher and student feature map. Results are shown in Table~\ref{tab:distance}. With a transformation from teacher to student, the distance is reduced before training. All the knowledge distillation methods show a reduction in KL divergence and $L_2$ distance, which implies that the student has learned knowledge from the teacher and becomes closer to the teacher. Specifically, our proposed method shows a considerable high consistency compared to those without the transformation operation. In other words, our proposed feature transformation promotes the original distillation methods obtaining more consistent features.

\section{Conclusion}
In this paper, we claim that students should be taught according to their fitness to the teacher while applying knowledge distillation methods.  
The proposed method provides an easier way for the student to learn from a transformed teacher feature. 
Different from existing methods, the internal discrepancy between student and teacher is considered and reduced. 
We have proved the promising results of the proposed method on different datasets and conducted plenty of ablation studies to verify the correctness of our intuition as well as the effect of the method.

{\small
\bibliographystyle{ieee_fullname}
\bibliography{egbib}
}

\end{document}